\documentclass[11pt]{article}

\usepackage[final]{acl}

\usepackage{times}
\usepackage{latexsym}

\usepackage[T1]{fontenc}

\usepackage[utf8]{inputenc}

\usepackage{microtype}

\usepackage{inconsolata}

\usepackage{graphicx}

\usepackage{booktabs}
\usepackage{amsmath}
\usepackage{makecell}
\usepackage{amssymb}
\usepackage{multirow}
\usepackage{booktabs}
\usepackage{array}
\usepackage{enumitem}

\title{MathEDU: Feedback Generation on Problem-Solving Processes for Mathematical Learning Support}

\author{Wei-Ling Hsu, Yu-Chien Tang, An-Zi Yen\\
       Department of Computer Science, National Yang Ming Chiao Tung University, Taiwan \\
      \texttt{weiling.hsu.cs11@nycu.edu.tw},
      \texttt{tommytyc.cs10@nycu.edu.tw},
      \texttt{azyen@nycu.edu.tw}}

\begin{document}
\maketitle
\begin{abstract}
The increasing reliance on Large Language Models (LLMs) across various domains extends to education, where students progressively use generative AI as a tool for learning. 
While prior work has examined LLMs' mathematical ability, their reliability in grading authentic student problem-solving processes and delivering effective feedback remains underexplored.
This study introduces MathEDU, a dataset consisting of student problem-solving processes in mathematics and corresponding teacher-written feedback.
We systematically evaluate the reliability of various models across three hierarchical tasks: answer correctness classification, error identification, and feedback generation.
Experimental results show that fine-tuning strategies effectively improve performance in classifying correctness and locating erroneous steps. 
However, the generated feedback across models shows a considerable gap from teacher-written feedback.
Critically, the generated feedback is often verbose and fails to provide targeted explanations for the student's underlying misconceptions.
This emphasizes the urgent need for trustworthy and pedagogy-aware AI feedback in education.
\end{abstract}

\section{Introduction}

The proliferation of generative AI systems, particularly LLMs, is rapidly reshaping the educational landscape, building on the trends of online learning that became mainstream in the post-pandemic era~\citep{alqahtani2020learning, jafar2022assessing,tack2022ai, kasneci2023chatgpt}. 
However, hallucinations pose serious risks by propagating misconceptions or introducing false content, and prior work shows that users are often persuaded by such fluent yet misleading outputs~\citep{mittelstadt2023protect}.  

\begin{figure}[t]
  \centering
  \includegraphics[width=\linewidth]{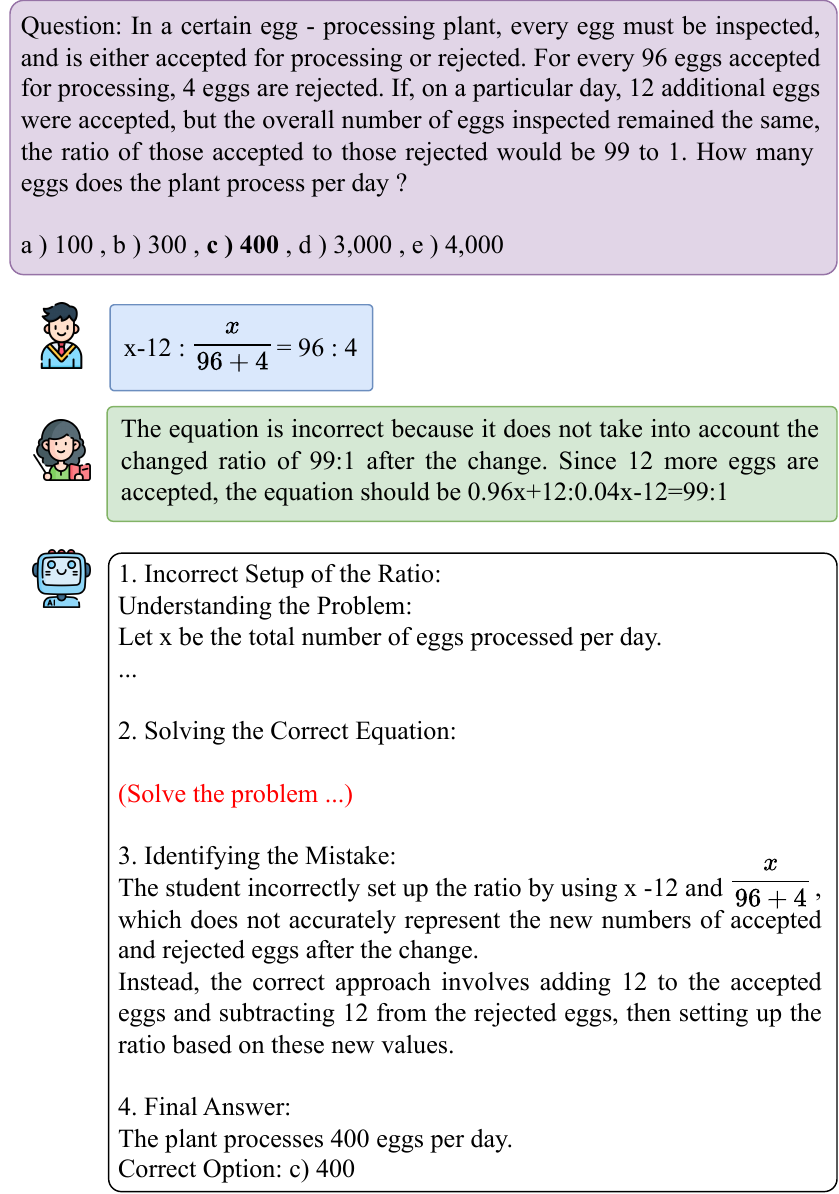}
  \caption{Example of Model Providing Lengthy but Inappropriate Feedback to the Student's Solution.}
  \label{fig:example}
\end{figure}

While early research demonstrated the potential of LLMs in solving mathematical word problems~\citep{shen2021generate, yu2021improving, jie2022learning}, these studies often focused on generating a correct final answer rather than evaluating the student's reasoning process. 
This overlooks a critical aspect of learning: understanding how a student arrives at a solution. 
A significant risk, as summarized by \citet{yen2023three}, is that LLMs can struggle to correctly interpret students' mathematical reasoning, sometimes even affirming a student's flawed logic. 
This raises a crucial question: How reliably can LLM/large reasoning model (LRM) assess the intermediate steps of a student's problem-solving process and provide feedback? 
To investigate this issue, we collect real student problem-solving processes alongside teacher-written feedback.
Figure~\ref{fig:example} presents a failure case using a real student response, which we evaluated with o1-mini~\cite{jaech2024openai}, an LRM, to determine correctness and generate feedback.
We found that the model ignored the student's actual reasoning process and overwhelmed them with extraneous detail that may increase cognitive load.
The generated explanation also failed to clearly pinpoint \textit{what was wrong}, \textit{why it was wrong}, or \textit{how to fix it}. 

\begin{table}[t]
    \centering
    \footnotesize
    \begin{tabular}{p{7.2cm}}
        \toprule
        \textbf{Problem:} Two trains of equal length are running on parallel lines in the same directions at 46 km/hr and 36 km/hr. The faster train passes the slower train in 144 seconds. The length of each train is: \\
        \textbf{Student Process}: 
        \[
        \begin{aligned} 
        & 46 - 36 = 10 \\
        & 10 \, \mathrm{km/hr} \times \frac{5}{18} = \frac{100}{36} \, \mathrm{m/s} \\
        & \frac{100}{36} \times 144 = 400 \, \mathrm{m} 
        \end{aligned}
        \] \\
        \midrule
        \textbf{Error Type:} Wrong Mathematical Operation/Concept \\
        \textbf{Error Equation:} 
        \[
        \frac{100}{36} \times 144 = 400 \, \mathrm{m}
        \] \\
        \textbf{Teacher Feedback:} To overtake the other train, you need to travel the combined length of both trains. Since both trains are of the same length, you need to divide by 2 to get the answer. \\
        \bottomrule
    \end{tabular}
    \setlength\tabcolsep{2pt}
    \caption{Example of Student Problem-Solving Process with Teacher Grading and Feedback.}    \label{tab:prompt_t_s}
\end{table}

To enable broader analysis, this work extends MathQA~\citep{amini2019mathqa} by adding annotations of students' problem-solving processes together with teacher feedback. 
In total, 4,048 annotated student solutions were collected, providing a resource for future research.
Table~\ref{tab:prompt_t_s} provides an example of the annotations, including error types, incorrect steps, and teacher feedback addressing misunderstandings.
We investigate how LLMs can evaluate students' reasoning, recognizing multiple valid approaches. 
Concretely, we conceptualize the task as three levels: (1) answer correctness classification, (2) error identification, and (3) feedback generation.
Answer correctness classification examines whether the model can identify whether the problem-solving process is correct.
Error identification assesses whether it can locate the specific error step.
Feedback generation is the most advanced level, requiring the model to transform that diagnosis into targeted and pedagogically actionable guidance.
In sum, our contributions are threefold:
(1)~This study investigates the challenge of assessing students' reasoning processes by exploring the use of LLMs in the context of mathematical education. 
(2)~We present the MathEDU dataset,\footnote{ \url{https://github.com/NYCU-NLP-Lab/MathEDU}} designed to address students' mistakes in math word problem-solving and provide personalized  feedback.
(3)~We evaluate LLM performance with prompting and LoRA~\cite{hu2021lora} fine-tuning.
The results show that fine-tuning enables the model to judge answer correctness more effectively, while generating feedback remains challenging.

\section{Related Work}

\subsection{Large Language Models for Education}

Several studies leverage LLMs to support instruction by generating distractors for multiple-choice questions~\citep{dave2021math}, analyzing students' answer histories~\citep{gao2021rcd}, and creating personalized instructional materials~\citep{he2024evaluating}. 
Others use LLMs to simulate students in interactive chats, enhancing classroom simulations for teacher training~\citep{markel2023gpteach}. 
LLMs have also been deployed as virtual tutors, such as generating code explanations in computer science classrooms~\citep{macneil2023experiences}.
However, 
LLMs often produce lengthy responses. 
In learning contexts, these overly detailed responses can cause information overload, reducing students' focus and diminishing active engagement~\cite{lo2024influence,zhang2025breaking}.

\subsection{Mathematics Education Datasets}

Several educational datasets have been developed, such as ASSISTments,\footnote{\url{https://www.etrialstestbed.org/resources/featured-studies/dataset-papers}}  Algebra 2005–2007,\footnote{\url{https://pslcdatashop.web.cmu.edu/DatasetInfo?datasetId=110}}\footnote{\url{https://pslcdatashop.web.cmu.edu/DatasetInfo?datasetId=330}} EdNet~\cite{choi2020ednet}, Junyi~\cite{JunyiOnlineLearningDataset}, and Eedi~\cite{wang2020instructions}, each capturing large-scale student interaction data across various subjects and grade levels. While these datasets provide valuable records for tasks like knowledge tracing, they primarily log final answers or coarse-grained interaction events, lacking detailed, step-by-step records of students' problem-solving processes. 
Teacher-written feedback on student errors is also absent from these studies.
MATHDIAL~\cite{macina2023mathdial} addresses this gap partially by using LLMs to simulate student responses and pairing them with teacher feedback.
\citet{daheim2024stepwise} constructed a dataset combining synthetic and limited annotated student responses to investigate step-level error detection and remediation. 
Similarly, \citet{gao2024llm} presented Math-Minos, leveraging automatically generated stepwise natural language feedback to train verifiers, yet their data relies on model outputs rather than genuine student work. 
However, prior work~\cite{aher2023using,markel2023gpteach} shows such simulated responses often fail to reflect realistic student reasoning. Thus, collecting real student problem-solving processes together with authentic teacher feedback remains essential for advancing research.
MathEDU is built upon authentic student problem-solving processes, each meticulously annotated with teacher-written feedback. 

\section{Dataset construction}

\begin{table}[t]
  \centering
  \small
    \begin{tabular}{p{7.2cm}}
    \toprule
    \textbf{Question:} A train running at the speed of 48 km/hr crosses a pole in 9 seconds. What is the length of the train? \\
    \textbf{Rationale:} Speed = (48 x 5 / 18) m/sec = (40 / 3) m/sec. length of the train = (speed x time). length of the train = (40 / 3 x 9) m = 120 m. answer is c. \\
    \textbf{Options:} a) 140 , b) 130 , c) 120 , d) 170 , e) 160 \\
    \textbf{Correct Option is:} C \\
    \bottomrule
    \end{tabular}%
  \caption{Example of a MathQA Instance.}
  \label{tab:mathqa_example}%
\end{table}%

\subsection{Problem-Solving Process Collection} 
We construct the MathEDU dataset based on MathQA.\footnote{The dataset selection criteria are detailed in Appendix~\ref{sec:data_selection}.}
Table~\ref{tab:mathqa_example} shows an example from MathQA. Each instance contains a math word problem, multiple-choice options, and a rationale, which is the reference solution providing detailed step-by-step reasoning for solving the problem. 
We removed the multiple-choice options, presenting the questions as open-ended tasks. 
To comprehensively assess students' mathematical abilities and collect problem-solving processes from individuals with varying skill levels, we recruited six students from different university departments. \footnote{The annotation guidelines are elaborated in Appendix~\ref{sec:Annotation Guidelines}.}\footnote{The question distribution and academic backgrounds of student annotators are detailed in  Appendix~\ref{sec:Student Majors}.}
To ensure that each student answered a balanced number of questions across all types, we sampled 4,500 unique questions from MathQA while preserving the proportional distribution of the original dataset. 
The six types include ``General,'' ``Gain,'' ``Physics,'' ``Geometry (Geo.),'' ``Probability (Prob.),'' and ``other.''
Each student was assigned 750 questions, with the distribution of question types consistent across students.
They wrote the problem-solving processes on paper, and their responses were then transcribed into \LaTeX{} format\footnote{\url{https://mathpix.com/}} to ensure machine readability.

\begin{figure*}[t]
  \centering
  \includegraphics[width=\linewidth]{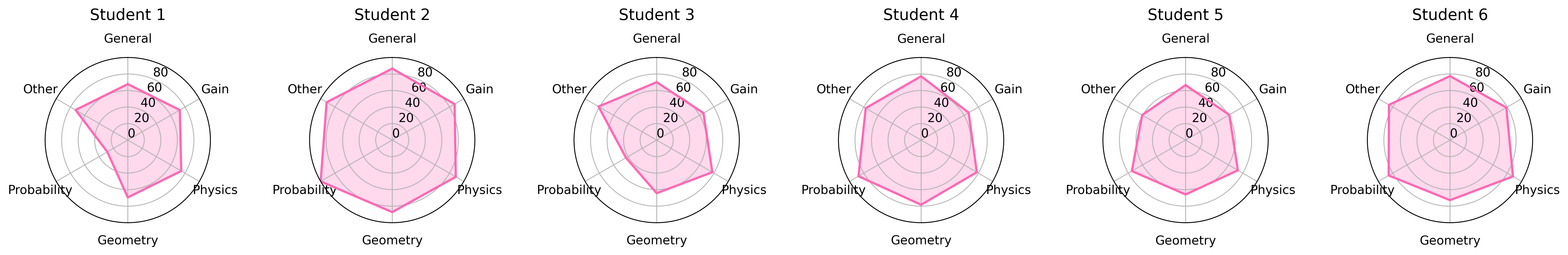}
  \caption{Performance of six students across six subject categories (General, Gain, Physics, Geometry, Probability, and Other). Scores are expressed as percentages, with higher values indicating better correctness.}
  \label{fig:radar}
\end{figure*}

\begin{figure}[t]
  \centering
  \includegraphics[width=\linewidth]{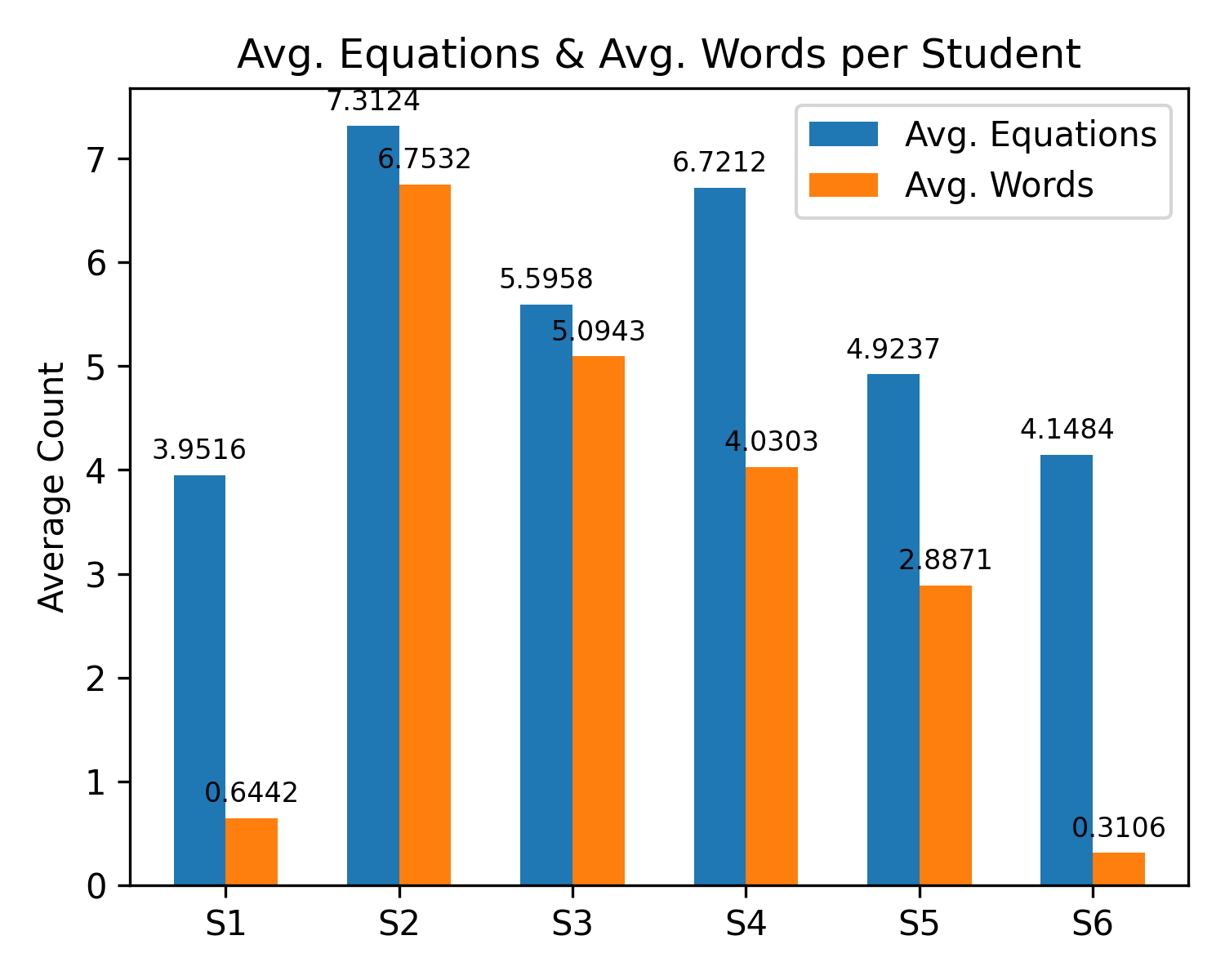}
  \caption{Average numbers of equations and words used by six students in their problem-solving processes.}
  \label{fig:bar}
\end{figure}

\subsection{Error and Feedback Annotation} 
After collecting the students' problem-solving results, we proceeded to label detailed grading information for each answer. 
We invited three experts in mathematics education to serve as teachers for grading and reviewing student responses. 
The three invited mathematics experts annotated the specific steps where errors occurred and providing explanations to clarify the reasons for these mistakes.
They also categorized the errors based on these annotated steps. 
Inspired by~\citet{wijaya2014difficulties}, We define three categories.
The error categories include ``Wrong Mathematical Operation/Concept'', ``Calculation Error'' and ``Incomplete Answer''. 
Additionally, to address instances where students either made careless mistakes or were completely unable to answer a question, we introduced two additional error categories: ``Careless Error'' and ``Lack of Necessary Mathematical Concepts''.\footnote{Descriptions and examples of these error categories are provided in Appendix~\ref{sec:Error Type Examples}.}

Each student's answer was annotated by three experts. 
In addition to labeling the types of errors and incorrect problem-solving steps, experts provided suggestions for students, including explanations for students' mistakes and suggestions for correcting them.
The questions are reviewed and removed erroneous ones that led to unrecognizable problem-solving processes, resulting in a dataset of 4,048 real student answers.\footnote{During the annotation process, we also asked the experts to flag any cases where the predefined categories did not fit the observed errors. But no such cases were reported.}
We measured the inter-rater agreement for error category annotations, 
using Krippendorff's Alpha. 
The result was 0.7818, indicating a substantial level of agreement among the experts in their annotations. 
When a disagreement occurs, we follow the majority rule to determine the final annotation.
In cases where a majority decision could not be reached, the instance was revisited by the three experts to reach a final consensus. 
Finally, we reviewed and refined the labeled data to ensure quality.
As a result, our dataset includes 3,050 correct answers and 998 incorrect ones.\footnote{The student-solution annotation process took about two months, and the expert annotation took approximately one and a half months.}

\subsection{Data Analysis}


Figure~\ref{fig:radar} visualizes students' problem-solving correctness across six subject categories. 
The radar charts highlight the diversity observed even among the six participants.
Correctness varied significantly, with the most proficient students achieving up to 87.59\%, while those with lower proficiency scored as low as 67.01\%. 
Furthermore, individual strengths differed across areas, with some students reaching 100\% accuracy in probability questions, whereas others struggled, achieving only 28.57\%.

We further analyze variations in students' problem-solving strategies. 
Since quantifying these strategies is challenging, we indirectly assess them by evaluating the number of mathematical expressions used by each student. 
Figure~\ref{fig:bar} shows the average use of equations and written explanations per problem, providing insights into their approaches. 
This analysis reveals that Students 1, 5, and 6 tend to have more concise problem-solving processes, while Students 2, 3, and 4 have more detailed and complex processes. 
Notably, Student 2 used an average of 7.3124 equations per problem, the highest among the group.

We also analyzed the average number of English words, excluding algebraic symbols and \LaTeX{} expressions, as students may provide verbal explanations.
The results reveal differences in text usage: Student 6 rarely used words, averaging 0.3106 per problem, while Student 2 provided detailed explanations, averaging 6.7532 words per problem.\footnote{More detailed statistics are provided in Appendix~\ref{sec:data_stat}.}

\section{Grading Problem-Solving Processes}

\subsection{Task Formulation}

In this section, we focus on how the tutor model is developed to analyze students' problem-solving processes. 
The $i$-th input data consists of the math word question $q_i$, the rationale $r_i$ provided in MathQA, and the student's answering process $s_i$, which includes both equations and written descriptions of their reasoning. 
Based on this input, the model performs three main tasks:

\noindent \textbf{Answer Correctness Classification:} 
This task determines whether the student's overall answer is correct, treated as a binary classification with the output label $c_i$.

\noindent \textbf{Problem-Solving Error Identification:} 
This task identifies incorrect equations in the process. 
The output $E_i = {e_{i,1}, e_{i,2}, \ldots, e_{i,\alpha}}$ consists of $\alpha$  identified incorrect equations.

\noindent \textbf{Feedback Generation:} 
This task generates feedback, including explanations or suggestions, to help students understand and correct their mistakes. 
The feedback is denoted as $T_i$. 



\subsection{Modeling Approach}
Three fine-tuning settings are explored.
In single-task training, separate models were fine-tuned for each of the three subtasks. 
In multi-task training, a single model was trained on data annotated for all three subtasks, using distinct prompts to guide the model in performing specific tasks. 
End-to-end training required the model to sequentially analyze the problem statement and evaluate the student's answer, providing a complete assessment without dividing the tasks explicitly.

Specifically, we employ LoRA fine-tuning to train an LLM  $\mathcal{M}$, parameterized by $\Phi$.
The training data used in LoRA fine-tuning is denoted as $\mathcal{Z} = \{(x_i, y_i)\}_{i=1, \ldots, N}$. 
Given task-specific parameter increment $\Delta\Phi = \Delta\Phi(\Theta)$, we optimize over $\Theta$:
\begin{equation}
    \label{eq:lora}
    \max_{\Theta} \sum_{(x,y) \in \mathcal{Z}} \sum_{t=1}^{|y|} (\log p_{{\Phi}_0 + \Delta\Phi(\Theta)}(y_t | x, y_{<t})).
\end{equation}

\noindent\textbf{Single-task and multi-task with rationale: }
In these two approaches, the training data used in LoRA fine-tuning is denoted as 
$
\mathcal{Z} = \{(x_i, c_i), (x_i, E_i), (x_i, T_i)\}_{i=1, \ldots, N},
$
where $x_i = \mathcal{P}_A(q_i, p_i, r_i)$. 
Here, $\mathcal{P}_A$ represents the prompt that takes $q_i$, $r_i$, and $p_i$ to generate grading results.

\noindent\textbf{Single-task and multi-task without rationale: }
Similar to the ``Single-task and multi-task with rationale'' approaches, these methods also use the same training approach but exclude the rationale in the input data. 
The training data used in LoRA fine-tuning is denoted as 
$
\mathcal{Z} = \{(x_i, c_i), (x_i, E_i), (x_i, T_i)\}_{i=1, \ldots, N},
$
where $x_i = \mathcal{P}_B(q_i, p_i)$. 
Here, $\mathcal{P}_B$ represents the prompt that takes $q_i$ and $p_i$ to generate grading results.

\noindent\textbf{End-to-end with rationale: }
In this approach, the training data used in LoRA fine-tuning is denoted as 
$
\mathcal{Z} = \{(x_i, y_i)\}_{i=1, \ldots, N},
$
where $x_i = \mathcal{P}_C(q_i, p_i, r_i)$ and $y_i = (c_i, E_i, T_i)$. 
Here, $\mathcal{P}_C$ represents the prompt that takes $q_i$, $r_i$, and $p_i$ to generate grading results.

\noindent\textbf{End-to-end without rationale: }
Similar to the previous approach, this method also uses an end-to-end training approach but excludes the rationale in the input data. 
The training data used in LoRA fine-tuning is denoted as 
$
\mathcal{Z} = \{(x_i, y_i)\}_{i=1, \ldots, N},
$
where $x_i = \mathcal{P}_D(q_i, p_i)$ and $y_i = (c_i, E_i, T_i)$. 
Here, $\mathcal{P}_D$ represents the prompt that takes $q_i$ and $p_i$ to generate grading results.

\begin{table*}[t]
  \centering
  \footnotesize
  \setlength\tabcolsep{2mm}
    \begin{tabular}{rlrrrrrrrrr}
    \toprule
          &       & \multicolumn{7}{c}{Answer Correctness Classification} & \multicolumn{2}{c}{ Error Identification} \\
          \cmidrule(lr){3-9}
          \cmidrule(lr){10-11}
          &       & \multicolumn{1}{c}{All} & \multicolumn{1}{c}{ General} & \multicolumn{1}{c}{Gain} & \multicolumn{1}{c}{ Physics} & \multicolumn{1}{c}{ Geo.} & \multicolumn{1}{c}{ Prob.} & \multicolumn{1}{c}{ Other} & \multicolumn{1}{c}{EM ($\uparrow$)} & \multicolumn{1}{c}{Dis. ($\downarrow$)} \\
    \midrule
    \multicolumn{1}{l}{Llama3 8B} & w/o $r$ & 66.96\% & 69.17\% & 67.23\% & 63.11\% & 80.00\% & 66.67\% & 56.41\% & 28.40\% & 67.61 \\
          & w/ $r$ & 77.04\% & 77.70\% & 74.07\% & 77.50\% & 84.62\% & 77.78\% & 69.77\% & 11.73\% & 92.71 \\
    \multicolumn{1}{l}{Llama3 70B} & w/o $r$ & 88.23\% & 87.10\% & 91.50\% & 86.49\% & 98.11\% & 58.82\% & 93.88\% & 32.10\% & 70.23 \\
          & w/ $r$ & 86.65\% & 83.72\% & 90.67\% & 86.06\% & \textbf{98.18\%} & 66.67\% & 89.80\% & 29.63\% & 70.66 \\
    \multicolumn{1}{l}{GPT-3.5} & w/o $r$ & 33.83\% & 28.27\% & 37.11\% & 38.37\% & 45.71\% & 42.86\% & 13.79\% & 30.25\% & 63.72 \\
          & w/ $r$ & 49.92\% & 45.79\% & 53.70\% & 54.45\% & 54.05\% & 30.77\% & 41.18\% & 27.78\% & 60.80 \\
    \multicolumn{1}{l}{o1-mini} & w/o $r$ & 94.66\% & \textbf{94.23\%} & 94.74\% & 95.06\% & 92.00\% & \textbf{94.12\%} & 97.96\% & \textbf{41.98\%} & 56.30 \\
    \midrule
    \multicolumn{1}{l}{Single-task } & w/o $r$ & 92.39\% & 90.49\% & 93.75\% & 93.66\% & 96.43\% & 81.82\% & 94.12\% & 24.69\% & 94.86 \\
          & w/ $r$ & \textbf{95.07\%} & 93.01\% & \textbf{96.86\%} & 95.68\% & 96.43\% & 90.00\% & \textbf{100.00\%} & 23.46\% & 96.43 \\
    \multicolumn{1}{l}{Multi-task } & w/o $r$ & 84.10\% & 83.15\% & 80.30\% & 89.16\% & 77.27\% & 71.43\% & 83.72\% & 23.46\% & 94.67 \\
          & w/ $r$ & 94.15\% & 92.46\% & 96.05\% & 94.66\% & 96.15\% & 88.89\% & 95.83\% & 17.28\% & 84.99 \\
    \multicolumn{1}{l}{End-to-end } & w/o $r$ & 92.09\% & 89.40\% & 95.71\% & 94.16\% & 93.10\% & 75.00\% & 94.12\% & 30.25\% & 71.01 \\
          & w/ $r$ & 94.38\% & 92.26\% & \textbf{96.86\%} & \textbf{96.09\%} & 96.43\% & 75.00\% & 98.04\% & 40.12\% & \textbf{52.60} \\
    \bottomrule
    \end{tabular}%
  \caption{Results of Answer Correctness Classification (Measured by F-score) and Problem-Solving Error Identification (Measured by Exact Match and Distance) Across Models and Settings.}
  \label{tab:exp_1_2}%
\end{table*}%

\section{Experiments}

\subsection{Experimental Setup} 

\noindent \textbf{Models.} We evaluated Llama3 8B~\citep{touvron2023Llama}, Llama3 70B, and GPT-3.5~\citep{ouyang2022training}\footnote{The version we used is \texttt{gpt-3.5-turbo-0125}.} using few-shot prompting.
We also explored the capabilities of the reasoning model (i.e., o1-mini~\citep{jaech2024openai}). 
We further fine-tuned Llama3 8B with a learning rate set to 2e-4 and a LoRA rank of 16.\footnote{All the prompts are shown at: \url{https://github.com/NYCU-NLP-Lab/MathEDU/blob/main/Prompts.pdf}.}
During inference, the temperature was set to 0, with a maximum output of 512 tokens, while other parameters remained at their default settings.

\noindent \textbf{Data Splitting.}
Student records were organized chronologically, with part of the records serving as the training set (answer history) and the remaining records as the validation and test sets in a 70:15:15 ratio. 
This setting assesses whether the model can leverage past responses to identify a student's strengths and weaknesses and apply this knowledge to grade new answers. 
Specifically, the training, validation, and test sets consist of 2,836, 609, and 603 examples, respectively. 

\noindent\textbf{Inference Prompt.}
For the few-shot prompting, six examples were provided, consisting of three correct and three incorrect student responses. 
We randomly selected the student's prior answer records from the training set as few-shot demonstrations.
For the fine-tuned model, which was extensively trained on the task, we employed zero-shot prompting to generate results directly.

\subsection{Experimental Results} 


\subsubsection{Answer Correctness Classification}
We evaluated the ability of LLMs to determine the correctness of student answers. 
F-score is adopted as the evaluation metric to address class imbalance, as correct answers are more frequent than incorrect ones.
The results are shown in Table~\ref{tab:exp_1_2}.\footnote{Note that the results of the two tasks are combined in a single table.
For answer correctness classification, ``All'' indicates the overall performance.}
We tested whether the detailed rationales in MathQA assist LLMs in grading student answers.
``w/o $r$'' and ``w/ $r$'' refer to whether the input includes the rationale of each question, respectively.
Llama3 70B achieved 88.23\% F-score, significantly outperforming Llama3 8B (66.96\%), indicating the importance of parameter size in mathematical computation and understanding. 
GPT-3.5's lower F-score stemmed from its strict evaluation of omitted calculations. 
For example, if a student's final answer was correct but some intermediate steps were skipped, GPT-3.5 often classified the answer as incorrect.
Notably, o1-mini demonstrates exceptional mathematical capabilities, achieving an overall F-score of 94.66\% in identifying the correctness of the student's answers. 
This represents the value of strong reasoning capabilities in correctness identification.\footnote{We did not conduct experiments with rationales, as they substantially increased input length, led to longer outputs, and reduced output stability.}

The bottom three rows of Table~\ref{tab:exp_1_2} represent our fine-tuned models, which demonstrate higher F-score in determining correctness compared to the original, non-fine-tuned Llama3 8B model. 
Among these, the end-to-end training method perform better than the multi-task training method. 
This improvement may be attributed to the end-to-end approach, which requires the model to generate grading results sequentially, enabling it to perform more coherent reasoning and achieve better outcomes.

The impact of including rationales in the input is evident in the results. 
For example, Llama3 8B's F-score rose from 66.96\% to 77.04\%, while Llama3 70B showed a slight decline.
Fine-tuned models demonstrated a stronger ability to effectively leverage rationales.
We further conducted McNemar's test to assess the significance of differences between our fine-tuned models and the baseline Llama3 70B without rationale. 
The single-task model with rationale significantly outperformed the baseline, with $p < 0.001$. 
Overall, leveraging LRMs can yield strong performance on this task, but for scenarios requiring a smaller-scale, more controllable system, fine-tuning a small LLM offers a cost-effective alternative that can even surpass the performance of much larger models.

\subsubsection{Problem-Solving Error Identification}

To evaluate the model's performance in identifying erroneous equations in cases where a student's answer is incorrect, we employ two metrics. 
The first is the exact match (EM) ratio, which measures the correspondence between model-detected step and manually annotated step. 
The second metric is ``Distance,'' computed using the Hausdorff distance between the model-detected and manually annotated error spans. This metric captures the deviation between predicted and gold spans, even when they do not overlap exactly.

The results are shown in Table~\ref{tab:exp_1_2}.
Non-fine-tuned LLMs achieved an EM ratio of around 30\% without rationale. 
However, providing rationale reduced the EM ratio for all models, with Llama3 8B showing the largest drop of 16.67\%. 
This may result from differing formats between the rationale and the student's actual process (e.g., \LaTeX{}-formatted), which increases the complexity of error identification.
For fine-tuned models, the multi-task model performed the worst without rationale. 
It rarely identified errors correctly, often gave irrelevant responses, and sometimes failed to output answers. 
Its performance further declined when rationale was included. The single-task model showed a similar pattern. 
In contrast, the end-to-end model outperformed Llama3 8B without rationale and showed noticeable improvements with rationale.
This result indicates that fine-tuning with this approach helps the model better understand and utilize rationale for error identification.

o1-mini demonstrated stronger mathematical reasoning and understanding abilities
, surpassing the end-to-end model. 
However, it performed worse in terms of the distance metric. 
Upon reviewing o1-mini's outputs, we found that in cases where errors were due to unfinished problems, the student's calculations were correct. 
Teachers typically did not label any formulas as incorrect. 
Despite this, o1-mini still misidentified incorrect formulas in these cases, which introduced significant penalties in the distance metric, leading to a poorer performance in this aspect.
This is a critical risk in educational settings, since over-diagnosing errors may unfairly penalize students and decrease trust in the system, and also reduce their willingness to learn.

\begin{table}[t]
  \centering
  \footnotesize
  \setlength\tabcolsep{2.5pt}
    \begin{tabular}{rlrrr}
    \toprule
          &       & \multicolumn{1}{l}{ROUGE-L} & \multicolumn{1}{l}{BERTScore} & \multicolumn{1}{l}{LLM Rating} \\
    \midrule
    \multicolumn{1}{l}{Llama3 8B} & w/o $r$ & 0.2391 & 0.7654 & 2.68 \\
          & w/ $r$ & 0.1632 & 0.5515 & 2.14 \\
    \multicolumn{1}{l}{Llama3 70B} & w/o $r$ & 0.2439 & 0.7147 & 3.61 \\
          & w/ $r$ & 0.2311 & 0.7108 & 3.31 \\
    \multicolumn{1}{l}{GPT-3.5} & w/o $r$ & 0.2695 & \textbf{0.8395} & 3.70 \\
          & w/ $r$ & 0.2597 & 0.8366 & 3.63 \\
    \multicolumn{1}{l}{o1-mini} & w/o $r$ & 0.1839 & 0.7986 & \textbf{4.70} \\
    \midrule
    \multicolumn{1}{l}{Single-task } & w/o $r$ & 0.1594 & 0.4836 & 0.63 \\
          & w/ $r$ & 0.0750 & 0.2182 & 0.39 \\
    \multicolumn{1}{l}{Multi-task } & w/o $r$ & 0.0124 & 0.0437 & 0.08 \\
          & w/ $r$ & 0.1552 & 0.4372 & 0.55 \\
    \multicolumn{1}{l}{End-to-end } & w/o $r$ & 0.1761 & 0.5327 & 0.99 \\
          & w/ $r$ & \textbf{0.2703} & 0.6786 & 1.91 \\
    \bottomrule
    \end{tabular}%
  \caption{Results of Feedback Generation.}
  \label{tab:teacher advic}%
\end{table}%

\subsubsection{Feedback Generation}
We evaluate the quality of these suggestions using three metrics: ROUGE-L ~\citep{lin2004rouge}, BERTScore~\citep{zhangbertscore}, and LLM Rating~\citep{liu2023g}. 
The LLM Rating, using GPT-4 as the evaluator, compares model-generated feedback against the ground-truth teacher suggestions and assigns a score from 0 to 5, where higher scores reflect better quality. 
Following established evaluation practices, GPT-4 is instructed to assess each response based on its relevance, helpfulness, and alignment with teacher feedback. A score of 0 indicates completely irrelevant or nonsensical suggestions, while a score of 5 reflects highly relevant, helpful, and well-aligned guidance. 
Table~\ref{tab:teacher advic} presents the results of feedback generation.\footnote{Additional results for ROUGE-1 and ROUGE-2 are provided in Appendix~\ref{app:rouge12}.} 
Larger models such as Llama3 70B and GPT-3.5 do not outperform Llama3 8B as expected, suggesting that generating teacher-like feedback remains a challenging task even for high-capacity models.

In our fine-tuned models, the single-task and multi-task models performed poorly, rarely identifying errors or generating feedback. 
Although the end-to-end model with rationale achieved ROUGE and BERTScore results comparable to Llama3 70B, they received lower LLM ratings. 
This represents the complexity of feedback generation, which requires larger models with stronger reasoning and language generation capabilities.
Even with fine-tuning, Llama3 8B appears to have reached its performance limit.
Specifically, the fine-tuned model excels at generating feedback for problem types it has been trained on, providing accurate suggestions. 
However, when confronted with unfamiliar problem types, its outputs tend to follow learned patterns, which may result in feedback that is less relevant or less accurate.

In the LLM rating conducted by using GPT-4, o1-mini received an almost perfect score of 4.70. 
However, its responses were often overly lengthy, generating additional irrelevant information, which impacted the evaluation results and further reduced their effectiveness in addressing specific student errors. 
Further discussion will be provided in the following section.

\begin{table}[t]
  \centering
  \setlength\tabcolsep{3mm}
  \resizebox{\linewidth}{!}{
    \begin{tabular}{lrrrrrr}
    \toprule
          & \multicolumn{1}{c}{All} & \multicolumn{1}{c}{Wrong M.} & \multicolumn{1}{c}{Calc.} & \multicolumn{1}{c}{Incomp.} & \multicolumn{1}{c}{Lack of M.} & \multicolumn{1}{c}{Careless} \\
    \midrule
    \multicolumn{1}{l}{Llama3 70B} & 1.61 & 1.59 & 1.30 & 1.33 & 2.20 & 2.00 \\
    \multicolumn{1}{l}{GPT-3.5} & 1.59 & 1.46 & 1.35 & \textbf{1.61} & 2.30 & \textbf{2.20} \\
    \multicolumn{1}{l}{o1-mini} & \textbf{1.82} & \textbf{1.79} & \textbf{1.65} & 1.44 & \textbf{2.55} & 1.60 \\
    \multicolumn{1}{l}{End-to-end} & 1.01 & 0.81 & 1.10 & 0.77 & 2.00 & 1.40 \\
    \bottomrule
    \end{tabular}%
    }
  \caption{Human Evaluation of Generated Feedback Across Error Categories.}
  \label{tab:human_evaluate}%
\end{table}%

\begin{figure*}[t]
  \centering
  \includegraphics[width=\linewidth]{figures/pies_1x4_globallegend_nooverlap.png}
  \caption{Feedback Categories by Models}
  \label{fig:pie}
\end{figure*}

\section{Discussion}

\subsection{Human Evaluation} 
\label{sec:human-eval}

To evaluate the model responses, we selected representative results from Llama3 70B without rationale, GPT-3.5 without rationale, and an end-to-end model with rationale. 
The o1-mini model, which uses few-shot prompts without rationale, is included.
The invited three mathematics experts were asked to evaluate the quality of the generated feedback.
The evaluation criteria range from 0 (not helpful) to 3 (clear and effective guidance). \footnote{The evaluation criteria is elaborated in Appendix~\ref{sec:eval_criteria}.}

The results are shown in Table~\ref{tab:human_evaluate}.
``Wrong M.,'' ``Calc.,'' ``Incomp.,'' ``Lack of M.,'' and ``Careless,'' correspond to ``Wrong Mathematical Operation/Concept,'' ``Calculation Error,'' ``Incomplete Answer,'' ``Lack of Necessary Mathematical Concepts,'' and ``Careless Error,'' respectively.
We assessed the inter-rater agreement on the models' output ratings using Krippendorff's Alpha, yielding a score of 0.7628, which reflects a moderate level of agreement among the experts.

The responses from language models remain below the ideal score of 3. 
All models excelled at handling ``Lack of Necessary Mathematical Concept'' errors, since these cases only require generating the correct solution and explanation.
By contrast, they struggled with ``Calculation Errors,'' often missing even simple mistakes, and sometimes misattributing errors or revising correct processes.
Our end-to-end model further lagged behind larger models on ``Wrong Operations/Concepts,'' ``Incomplete,'' and ``Careless Errors,'' reflecting its limited reasoning and process analysis capabilities.

Although GPT-3.5, o1-mini, and Llama3 70B have stronger reasoning and generation capabilities, they struggle to identify student errors and provide focused feedback. 
o1-mini often generates lengthy but inaccurate responses, while GPT-3.5 tends to solve problems directly rather than analyze student reasoning.
In contrast, our end-to-end model offers concise, targeted feedback but struggles with unfamiliar problem types.
These findings highlight the need for refining LLMs to better address the nuances of student problem-solving processes.
Besides, experts noted that Llama3 70B occasionally produced feedback with a judgmental tone, such as ``\textit{The student doesn't know how to set up the equation}.''
Such phrasing, while factually descriptive, risks discouraging students rather than guiding improvement, underscoring the importance of pedagogical sensitivity in feedback generation.
Additional case studies of feedback generation are provided in Appendix~\ref{sec:Example of Feedback Provided by the Model}.

\begin{table}[t]
\centering
\scriptsize
\setlength{\tabcolsep}{1mm}
\resizebox{\linewidth}{!}{
\begin{tabular}{lcccccc}
\toprule
\textbf{Model} & \textbf{S1} & \textbf{S2} & \textbf{S3} & \textbf{S4} & \textbf{S5} & \textbf{S6} \\
\midrule
Llama3 8B w/o $r$ 
& 59.59\% & 70.36\% & 61.50\% & 67.88\% & 66.72\% & 55.15\% \\
Llama3 8B w/ $r$  
& 62.37\% & 72.26\% & 64.90\% & 71.06\% & 73.90\% & 66.97\% \\
End-to-end w/o $r$
& 87.12\% & 57.96\% & 85.55\% & 87.73\% & 90.03\% & 89.09\% \\
End-to-end w/ $r$
& 94.58\% & 95.33\% & 90.86\% & 89.85\% & 96.48\% & 91.67\% \\
\midrule
Average
& 75.92\% & 73.98\% & 75.70\% & 79.13\% & 81.78\% & 75.72\% \\
\bottomrule
\end{tabular}
}
\caption{Accuracy of Correctness Classification Across Students (S1--S6).}
\label{tab:correctness_accuracy_by_student}
\end{table}

\begin{table*}[t]
\centering
\scriptsize
\setlength{\tabcolsep}{1.5mm}
\begin{tabular}{lrrrrrrrrrrrr}
\toprule
\multirow{2}{*}{\textbf{Model}} &
\multicolumn{2}{c}{\textbf{S1}} & \multicolumn{2}{c}{\textbf{S2}} &
\multicolumn{2}{c}{\textbf{S3}} & \multicolumn{2}{c}{\textbf{S4}} &
\multicolumn{2}{c}{\textbf{S5}} & \multicolumn{2}{c}{\textbf{S6}} \\
\cmidrule(lr){2-3}\cmidrule(lr){4-5}\cmidrule(lr){6-7}\cmidrule(lr){8-9}\cmidrule(lr){10-11}\cmidrule(lr){12-13}
& \textbf{EM ($\uparrow$)} & \textbf{Dis. ($\downarrow$)} & \textbf{EM ($\uparrow$)} & \textbf{Dis. ($\downarrow$)} & \textbf{EM ($\uparrow$)} & \textbf{Dis. ($\downarrow$)}
& \textbf{EM ($\uparrow$)} & \textbf{Dis. ($\downarrow$)} & \textbf{EM ($\uparrow$)} & \textbf{Dis. ($\downarrow$)} & \textbf{EM ($\uparrow$)} & \textbf{Dis. ($\downarrow$)} \\
\midrule
Llama3 8B w/o $r$
& 15.42\% & 48.69  & 3.53\%  & 174.22 & 7.65\%  & 137.87
& 4.91\%  & 140.47 & 13.33\% & 66.20  & 18.60\% & 37.30 \\
Llama3 8B w/ $r$
& 11.44\% & 50.36  & 8.24\%  & 187.19 & 4.59\%  & 136.57
& 0.61\%  & 150.21 & 9.33\%  & 73.39  & 18.60\% & 41.12 \\
End-to-end w/o $r$
& 38.81\% & 40.51  & 2.35\%  & 199.51 & 20.92\% & 117.27
& 20.25\% & 90.90  & 41.33\% & 46.28  & 52.71\% & 27.14 \\
End-to-end w/ $r$
& 58.71\% & 23.53  & 22.35\% & 96.47  & 28.57\% & 97.27
& 30.06\% & 92.64  & 61.33\% & 20.88  & 55.81\% & 24.50 \\
\midrule
Average
& 31.10\% & 40.77
& 9.12\%  & 164.35
& 15.43\% & 122.25
& 13.96\% & 118.55
& 31.33\% & 51.69
& 36.43\% & 32.51 \\
\bottomrule
\end{tabular}
\caption{Model Performance on Error Identification Across Students (Exact Match and Hausdorff Distance).}
\label{tab:em_dis_by_student}
\end{table*}

\begin{table*}[t]
\centering
\scriptsize
\setlength{\tabcolsep}{4pt}
\begin{tabular}{lcccccccccccc}
\toprule
\multirow{2}{*}{\textbf{Model}} &
\multicolumn{2}{c}{\textbf{S1}} & \multicolumn{2}{c}{\textbf{S2}} &
\multicolumn{2}{c}{\textbf{S3}} & \multicolumn{2}{c}{\textbf{S4}} &
\multicolumn{2}{c}{\textbf{S5}} & \multicolumn{2}{c}{\textbf{S6}} \\
\cmidrule(lr){2-3}\cmidrule(lr){4-5}\cmidrule(lr){6-7}\cmidrule(lr){8-9}\cmidrule(lr){10-11}\cmidrule(lr){12-13}
& \textbf{R-L} & \textbf{Rating} & \textbf{R-L} & \textbf{Rating}
& \textbf{R-L} & \textbf{Rating} & \textbf{R-L} & \textbf{Rating}
& \textbf{R-L} & \textbf{Rating} & \textbf{R-L} & \textbf{Rating} \\
\midrule
Llama3 8B w/o $r$
& 0.2269 & 2.98 & 0.1967 & 2.95 & 0.1972 & 2.99
& 0.1689 & 2.87 & 0.1945 & 2.35 & 0.1701 & 2.14 \\
Llama3 8B w/ $r$
& 0.1786 & 3.00 & 0.1459 & 2.86 & 0.1807 & 3.06
& 0.1372 & 2.67 & 0.1655 & 2.55 & 0.1487 & 1.97 \\
End-to-end w/o $r$
& 0.1738 & 2.16 & 0.0425 & 1.68 & 0.1615 & 2.08
& 0.2165 & 2.64 & 0.2040 & 1.66 & 0.2339 & 1.51 \\
End-to-end w/ $r$
& 0.2885 & 2.60 & 0.2676 & 2.51 & 0.2557 & 2.63
& 0.2039 & 3.00 & 0.3226 & 2.67 & 0.2301 & 2.00 \\
\midrule
Average
& 0.2169 & 2.69
& 0.1632 & 2.50
& 0.1988 & 2.69
& 0.1816 & 2.80
& 0.2216 & 2.31
& 0.1957 & 1.91 \\
\bottomrule
\end{tabular}
\caption{Model Performance on Feedback Generation Across Students (ROUGE-L and LLM Rating).}
\label{tab:rougel_rating_by_student}
\end{table*}

\subsection{Generated Feedback Characterization}

Following Section~\ref{sec:human-eval}, we extended the analysis of feedback generation with an automated, LLM-assisted approach to better capture pedagogical risks.
Given the scale of samples and models, fully manual annotation was impractical.
Inspired by \citet{chen2025symbolic}, we employed GPT-4o\footnote{The version used is \texttt{gpt-4o-2024-11-20}.} as a judge, evaluating each instance five times and consolidating results to ensure stability.
Finally, all final results were manually reviewed.
We categorized model-generated feedback errors into the following types:  
(1)~\textbf{Aligned}: Feedback closely matches teacher-written feedback.  
(2)~\textbf{Redundant detail}: Includes excessive, non-essential content that reduces focus.  
(3)~\textbf{Imprecise wording}: Uses inaccurate or inconsistent terminology (e.g., confusing ``area'' with ``volume''), creating ambiguity for students.  
(4)~\textbf{Missing steps}: Omits one or more essential reasoning steps.  
(5)~\textbf{Calculation error}: Produces incorrect numerical results. 
(6)~\textbf{Misjudged answer}: Incorrectly classifies the student's final answer as right or wrong.  
(7)~\textbf{Misidentified error}: Correct feedback but pointing to the wrong error location.  
(8)~\textbf{Misunderstood approach}: Misinterprets the student's problem-solving strategy. 

The results are shown in Figure~\ref{fig:pie}.
Overall, alignment with teacher feedback was rare; o1-mini often produced lengthy but less error-prone responses; GPT-3.5 and Llama3 70B often skipped steps or made calculation mistakes; our end-to-end model struggled with precise wording and misjudgments, indicating weaker generalization. All models exhibited reliability issues, such as misidentifying errors or misunderstanding student reasoning.

\section{Impact of Student Proficiency on Model Performance}

To address the variability across individual students, we report model performance on answer correctness classification, error identification, and feedback generation for each of the six students in Tables~\ref{tab:correctness_accuracy_by_student}, \ref{tab:em_dis_by_student}, and \ref{tab:rougel_rating_by_student}, respectively.
We focus on Llama3 8B and our end-to-end fine-tuned model, as they represent the baseline few-shot prompting approach and our best-performing fine-tuning strategy, allowing for a direct comparison between these two paradigms.
Based on overall correctness rates reported in Table~\ref{tab:student_record}, we categorize students into three proficiency levels: high-performing (S2: 87.59\%, S6: 80.61\%), medium-performing (S4: 75.30\%, S3: 71.09\%, S1: 70.57\%), and lower-performing (S5: 67.01\%).

Model performance does not exhibit a clear correlation with student proficiency levels. S2, despite being the highest-performing student, yields the lowest model performance in error identification. 
However, S6, also a high-performing student, achieves relatively better error identification results but receives the lowest LLM ratings for feedback quality.
Meanwhile, S5 obtains the highest accuracy in correctness classification, though this may simply reflect that certain types of errors are more straightforward to detect, rather than indicating superior model capability.
We hypothesize this disparity stems from problem-solving complexity. 
As shown in Figure~\ref{fig:bar}, S2 employs the most equations and words per problem, 
which may account for the poor error identification performance. 
S6's concise style with minimal verbal explanation may facilitate error localization but simultaneously limits the contextual information available for generating feedback.
These findings suggest that different tasks may be influenced by different aspects of student problem-solving behavior.

\section{Conclusion}

This study focuses on the application of generative AI in mathematics education, particularly its reliability in grading authentic student problem-solving processes and generating feedback. 
To facilitate a rigorous analysis, we construct the MathEDU dataset, which includes real student problem-solving processes and teacher feedback for GRE-level math problems. 
A range of models is evaluated on answer correctness classification, error identification, and feedback generation. 
Experimental results show that our fine-tuned models achieve notable improvements in classifying correctness and identifying erroneous steps, while LRMs demonstrate stronger overall performance across the three tasks. 
Nevertheless, current models still struggle with generating personalized suggestions. 
Common issues include imprecise wording, numerical miscalculations, omission of critical intermediate steps, and lengthy yet untargeted responses.
An advanced method is required to build pedagogy-aware AI systems that can both produce reliable, adaptive guidance and better interpret students' mathematical reasoning, which we leave as future work.
Additionally, extending the scale and scope to a broader range of problem types and domains is left as our future work.

\appendix

\section*{Limitations}


\noindent\textbf{Lack of Student Evaluation. }
In the current evaluation process, the original students are not involved in reviewing the grading results generated by the LLMs, with only peer students providing feedback. 
This lack of direct student input may lead to the loss of the most relevant insights. 
Additionally, since expert annotators evaluating the grading results do not interact with the students, it is difficult to determine to what extent the teacher's feedback helps students understand their mistakes or improve their learning. 
In future work, we plan to gather direct opinions from students to determine whether the feedback generated by the model are genuinely helpful to them. 
By involving students in responding to the model's grading and recommendations, we can better understand whether the feedback effectively helps students comprehend and correct their mistakes. 
Such feedback will not only improve the model's performance but also enhance its applicability in educational contexts.

\noindent\textbf{Limited to Mathematics. }
While we have conducted several experiments to discuss and analyze the capabilities of LLMs in mathematics tutoring, evaluating their potential advantages and disadvantages in real-world scenarios, our study is constrained to the mathematics domain due to the limitations of our dataset. 
Research in other fields remains significantly underexplored. 
In the future, we plan to explore the pedagogical capabilities of LLMs in different educational areas, such as programming instruction, to further uncover their potential in diverse subjects.


\noindent\textbf{Limited Dataset Size. }
The dataset is constrained to 4,048 entries due to the complexity and effort required for detailed data annotation, limiting its size and potentially reducing its effectiveness in  fine-tuning smaller models. 
Despite its limitations, the dataset offers authentic student problem-solving processes at the GRE difficulty level, accompanied by detailed feedback from expert teachers.
It is expected that this resource can support future research in mathematics education, particularly in advancing automated assessment of student answers.

\noindent\textbf{Limited Model and Method Exploration.}
Our study focused on Llama3 8B, Llama3 70B, and GPT-3.5, with experiments limited to few-shot prompting and LoRA-based fine-tuning. 
More recent LLMs remain unexplored, as keeping pace with rapidly evolving releases and conducting reliable expert-based evaluations poses substantial challenges. To ensure consistency and completeness, we focused on a representative set of models in this study.
Additionally, our fine-tuning methods did not delve into incorporating students' individual learning progress, such as tracking their mastered concepts.
Despite these limitations, we hope our work can contribute to the growing exploration of LLM applications in mathematics education and encourage further research to refine and expand these methods.

\section*{Ethics Statement}
To address potential privacy concerns associated with the MathEDU dataset, this section outlines the measures taken to protect annotators' privacy and the ethical considerations in releasing the dataset.
The MathEDU dataset contains problem-solving records annotated by students from diverse academic backgrounds. 
However, no personally identifiable information, such as names or IDs, was collected. 
Each record is assigned an anonymized identifier to categorize entries from the same student. While the dataset includes the students' academic backgrounds (i.e., majors), this information is provided solely for analytical purposes.
As highlighted in previous sections, understanding how problem-solving styles vary across mathematical abilities and expertise is a valuable area of research.
All annotators involved in the task, including students and mathematics experts serving as teachers, were compensated based on the minimum hourly wage standards of their respective countries.
The student annotators were informed that their personal data would remain confidential and that only their academic background would be disclosed. 
This consent was obtained prior to participation in the annotation task. 
The dataset's structure ensures that individual privacy is preserved while facilitating research in educational applications.
In alignment with ethical guidelines, the dataset will only be released for research purposes.

\section*{Acknowledgments}

This research was partially supported by National Science and Technology Council, Taiwan, under grant NSTC 114-2221-E-A49-057-MY3.

\bibliography{custom}

\appendix

\section{Math Word Problem Dataset Selection} \label{sec:data_selection}

Existing open datasets in mathematical domains include datasets such as MAWPS~\cite{koncel2016mawps}, Math23k~\cite{wang2017deep},  SVAMP~\cite{patel2021nlp}, and MathQA~\citep{amini2019mathqa}. 
These datasets provide numerous mathematical problems and reference rationales. 
To gather high-quality student processes, we have two main requirements for the mathematical problems in our dataset:

\begin{enumerate}[nolistsep]
    \item \textbf{Problem Difficulty:} We aim for the dataset to include both problems that can be solved with brief processes and those that are challenging and require complex solution paths. 
    This varying levels of difficulty allow us to obtain diverse problem-solving approaches and explore effective methods for assessing the correctness of these differing strategies. 
    \item \textbf{Problem Diversity:} We seek a dataset which has multiple domains such as probability, geometry, and algebra. 
    This diversity enables us to comprehensively assess students' mathematical abilities and further explore techniques applicable to grading problems across different domains. 
\end{enumerate}

Considering these factors, we utilize the MathQA dataset~\citep{amini2019mathqa}. 
MathQA includes problems from six distinct domains, covering a range of difficulties suitable for GRE-level problems.
Furthermore, MathQA not only provides answers to the problems but also includes rationales, which are correct solution processes that aid teachers in evaluating student process.

\begin{table*}[t]
\centering
\scriptsize
\begin{tabular}{p{15cm}}
\toprule
\textbf{Goal:} \\
The goal is to obtain students' problem-solving processes when they make errors in solving math problems, which will help train language models for applications in math education. \\ \\
\textbf{Annotation Process:} \\
A Google Sheet link will be provided to the students, which will contain both the Chinese and English versions of the questions. Students can choose one of the following methods to complete their solutions: \\
\begin{itemize}
  \item 1: Solve on a tablet.
  \item 2: Solve on paper and take a picture afterward (one picture per question). After completing the solutions, students should name the file as \texttt{<problem\_id>}  (e.g., \texttt{2456.jpg}) and upload it to the cloud drive.
\end{itemize} \\
The timeline for completing 750 questions will be discussed with students to set a deadline (estimated to be within two months). Weekly meetings will be scheduled to report progress and understand the students' problem-solving status. There will be no weekly minimum requirement; students can allocate their time as they see fit. \\ \\
\textbf{Annotation Guidelines:} \\ 
All problems are junior high school-level math problems. Students must show the full solution process and the final answer with the following requirements: \\
- Clearly write the solution (top-down). \\
- Include simple English explanations, no Chinese. \\
- Example 1 : \\
\[
\begin{aligned}
    & \text{Assume lent } X \\
    & X \times 0.08 \times 8 = 0.4X \\
    & 0.6X = 480 \\
    & X = 800
\end{aligned}
\]
\\
- Example 2 : \\
\[
\begin{aligned}
    & \text{Thief speed}: 50 \ \text{km/hr} \\
    & \text{Owner speed}: 60 \ \text{km/hr} \\
    & 50 \times 0.5 = 25 \ \text{km} \\
    & \text{The thief drove 25 km in half an hour.} \\
    & \frac{25}{60 - 50} = 2.5
\end{aligned}
\]
\\ 
- Round to 2 decimal places. Use 3.14 for pi. Simplify fractions. No calculators allowed.  \\
- If you encounter a question you cannot solve, note the problem number and the type of error, then skip it. Error types include: \\
\begin{itemize}
  \item 1: Don't know how to solve it.
  \item 2: Problem definition is unclear.
  \item 3: Unrecognizable symbols or notation.
  \item 4: Problem requires a diagram, but none is provided.
  \item 5: Other...
\end{itemize} \\

- If you feel the instructions are unclear or have any other questions, please ask the responsible staff immediately. \\

\bottomrule
\end{tabular}
\caption{Math Problem-Solving Process Guidelines.}
\label{tab:guideline}
\end{table*}

\section{Supplementary Details for Dataset Construction}\label{sec:Annotation}

\subsection{Annotation Guidelines }\label{sec:Annotation Guidelines}

Table~\ref{tab:guideline} presents the guidelines we used for collecting students' math word problem solving results.
The guidelines begin by explaining the rationale for annotating answers and providing instructions for submitting their work, whether by solving problems on a tablet or solving on paper and uploading a photo. 
We emphasize the importance of clear annotations and encourage students to include explanations in their problem-solving process. 
Additionally, we provide examples of proper annotation, and finally, we give instructions on how to handle problems they find unclear.

\subsection{Academic Backgrounds and Question Distribution of Student Annotators}\label{sec:Student Majors}

To collect data on math problem-solving, we invited six university students to participate in the annotation process.
Table~\ref{tab:student_question_distribution} presents the majors of the student annotators and the distribution of questions they answered.
We aimed for diversity by including individuals from different academic backgrounds, such as Japanese Studies and Applied Mathematics, to capture a range of math proficiency levels in the problem-solving process.
Before the annotation process, students were asked to attempt 10 questions to ensure their responses adhered to the required format. 
This step helped avoid issues such as incomplete equations or disorganized layouts, facilitating smoother data collection.

\begin{table*}[t]
  \centering
  \small
  \setlength\tabcolsep{2.2pt}
  \begin{tabular}{lccccccccc}
    \toprule
    ID & Major & Total & General & Gain & Physics & Geometry & Probability & Other \\
    \midrule
    S1 & Applied Mathematics & 683 & 276 (40.41\%) & 132 (19.33\%) & 182 (26.64\%) & 49 (7.17\%) & 7 (1.02\%) & 37 (5.42\%) \\
    S2 & Finance & 685 & 297 (43.36\%) & 133 (19.42\%) & 188 (27.45\%) & 39 (5.69\%) & 4 (0.58\%) & 24 (3.50\%) \\
    S3 & Japanese & 678 & 281 (41.45\%) & 136 (20.06\%) & 180 (26.55\%) & 42 (6.19\%) & 7 (1.03\%) & 32 (4.72\%) \\
    S4 & Information Management & 660 & 268 (40.60\%) & 140 (21.21\%) & 167 (25.30\%) & 46 (6.97\%) & 8 (1.21\%) & 31 (4.70\%) \\
    S5 & Mathematics Education & 682 & 275 (40.32\%) & 139 (20.38\%) & 186 (27.27\%) & 41 (6.01\%) & 8 (1.17\%) & 33 (4.84\%) \\
    S6 & Physics & 660 & 262 (39.70\%) & 127 (19.24\%) & 175 (26.52\%) & 55 (8.33\%) & 7 (1.06\%) & 34 (5.15\%) \\
    \bottomrule
  \end{tabular}

  \caption{Student Majors and Number of Questions Answered.}\label{tab:student_question_distribution}
\end{table*}

\section{Supplementary Information for Each Error Category }\label{sec:Error Type Examples}

The detailed definitions of these five error categories are as follows:
\textbf{(1)~Wrong Mathematical Operation/Concept} refers to the use of incorrect operations or inappropriate concepts. \textbf{(2)~Calculation Error} involves mistakes in arithmetic, equation solving, or unit conversion. \textbf{(3)~Incomplete Answer} occurs when a student begins a correct procedure but does not finish it. \textbf{(4)~Careless Error} captures inattentive mistakes, such as incorrect number substitution or missing digits. \textbf{(5)~Lack of Necessary Mathematical Concepts} indicates errors stemming from insufficient foundational knowledge.
While both (1) and (5) involve conceptual issues, the former reflects misapplication, whereas the latter involves an inability to begin due to missing knowledge.
The following are examples of each error category.

\noindent \textbf{Wrong Mathematical Operation/Concept:} 
This category refers to errors where the student applies an incorrect mathematical operation or employs an inappropriate mathematical concept when attempting to solve a problem. 
Such mistakes often occur when students misunderstand the nature of the mathematical task or misinterpret key elements of the problem. 
These errors can stem from a variety of issues, including a failure to correctly identify the mathematical procedure required or the application of an irrelevant concept to the problem at hand.
Additionally, this category encompasses cases where the student misinterprets critical keywords or phrases, leading to the incorrect selection of operations.
For example, misunderstanding terms such as ``per,'' ``rate,'' or ``difference'' can result in choosing the wrong formula or approach. 
Furthermore, errors in selecting or using relevant information from the problem, either by focusing on irrelevant data or neglecting essential variables, fall under this category.
Such errors reflect deeper issues in comprehension or the application of mathematical principles. 
By recognizing and addressing these types of mistakes, it becomes possible to better understand the student's reasoning process and to provide more targeted feedback for improving their problem-solving abilities.

In Table~\ref{tab:example_wrong_math}, we provide an example of this type of error. 
The problem involves two trains of the same length traveling in the same direction, with the speeds of each train and the time it takes for the faster train to catch up to the slower one provided. 
The question asks for the length of the trains. 
While the student correctly calculated that the distance covered by the faster train during the overtaking process was 400 meters, they failed to account for the fact that the faster train needs to cover the combined length of both trains to complete the overtaking maneuver, which led to the incorrect answer.

\begin{table}[t]
    \centering
    \footnotesize
    \begin{tabular}{p{7.2cm}}
        \toprule
        \textbf{Problem:} Two trains of equal length are running on parallel lines in the same directions at 46 km/hr and 36 km/hr. The faster train passes the slower train in 144 seconds. The length of each train is: \\
        \textbf{Student Process}: 
        \[
        \begin{aligned} 
        & 46 - 36 = 10 \\
        & 10 \, \mathrm{km/hr} \times \frac{5}{18} = \frac{100}{36} \, \mathrm{m/s} \\
        & \frac{100}{36} \times 144 = 400 \, \mathrm{m} 
        \end{aligned}
        \] \\
        \hline
        \textbf{Error Type:} Wrong Mathematical Operation/Concept \\
        \textbf{Error Equation:} 
        \[
        \frac{100}{36} \times 144 = 400 \, \mathrm{m}
        \] \\
        \textbf{Teacher Feedback:} To overtake the other train, you need to travel the combined length of both trains. Since both trains are of the same length, you need to divide by 2 to get the answer. \\
        \bottomrule
    \end{tabular}
    \setlength\tabcolsep{2pt}
    \caption{Example of Wrong Mathematical Operation/Concept}
    \label{tab:example_wrong_math}
\end{table}

\noindent\textbf{Calculation Error:} 
This category captures mistakes made by students during the computational phases of problem-solving, where the execution of arithmetic or algebraic operations is incorrect. 
Such errors can range from simple miscalculations in basic arithmetic to more complex mistakes involving the manipulation of algebraic expressions or functions. 
These errors do not typically stem from a misunderstanding of the problem's structure or concept but rather from faulty execution during the calculation process.
Table~\ref{tab:example_calculation} reports an example of this type of error.
The student made a calculation mistake when evaluating \(\frac{825 - 750}{5}\), resulting in an answer of 25 instead of the correct value of 15, leading to an incorrect answer for the problem.

\begin{table}[t]
    \centering
    \footnotesize
    \setlength\tabcolsep{2pt}
    \begin{tabular}{p{7.2cm}}
        \toprule
        \textbf{Problem:} at what rate percent on simple interest will rs . 750 amount to rs . 825 in 5 years ?\\
        \textbf{Student Process}: 
        \[
        \begin{aligned} 
        & \frac{825 - 750}{5} = 25 \\ 
        & \frac{25}{750} = \frac{1}{30} \approx 0.03333 \text{ (or } 3.33\% \text{)} 
        \end{aligned}
        \] 
        \\
        \hline
        \textbf{Error Type:} Calculation Error \\
        \textbf{Error Equation:} 
        \[
        \frac{825 - 750}{5} = 25
        \]
        \textbf{Teacher Feedback:} Incorrect calculation. The correct answer is 15. \\
        \bottomrule
    \end{tabular}
    \caption{Example of Calculation Error}
    \label{tab:example_calculation}
\end{table}

\noindent\textbf{Incomplete Answer:} 
The ``Incomplete Answer'' category refers to instances where the student begins solving a problem using the correct formula or procedure but does not carry the solution through to completion. 
Although the initial steps of the problem-solving process may be accurate and aligned with the required methodology, the student ultimately halts the process prematurely. 
This may result from a partial understanding of the problem or a lack of familiarity with subsequent steps needed to finalize the solution.
In Table~\ref{tab:example_unfinish}, an example of this error is illustrated. The problem asks for the average number of apples sold per hour over a two-hour period. However, the student calculated only the total number of apples sold during those two hours and stopped there, failing to continue to calculate the average, which led to an incorrect answer.

\begin{table}[t]
    \centering
    \footnotesize
    \setlength\tabcolsep{2pt}
    \begin{tabular}{p{7.2cm}}
        \toprule
        \textbf{Problem:} maria sold 10 kg of apples in her first hour at the market , but only 2 kg of apples in the second hour . on average , how many kg of apples did she sell in two hours at the market ? \\
        \textbf{Student Process}: 10+2=12 \\
        \hline
        \textbf{Error Type:} Incomplete Answer \\
        \textbf{Error Equation:} 
        None \\
        \textbf{Teacher Feedback:}
        After calculating the total number of apples sold in two hours, you still need to divide by the time to get the average sales per hour. Therefore, the answer is 12/2 = 6. \\
        \bottomrule
    \end{tabular}
    \caption{Example of Incomplete Answer}
    \label{tab:example_unfinish}
\end{table}

\noindent\textbf{Careless Error:} 
The ``Careless Error'' category encompasses mistakes that arise not from a misunderstanding of the mathematical concepts or procedures, but rather from inattentiveness or lapses in concentration during the problem-solving process. 
These errors often result from avoidable mistakes, such as substituting incorrect numbers into a formula or writing the wrong number. 
Table~\ref{tab:example_careless}, shows an example of this type of error. 
The problem provides the number 78; however, the student mistakenly interprets this number as 70 during the calculations, leading to an incorrect result.

\begin{table}[t]
    \centering
    \footnotesize
    \setlength\tabcolsep{3pt}
    \begin{tabular}{p{7.2cm}}
        \toprule
        \textbf{Problem:} Peter's average (arithmetic mean) test score on 4 tests is 78. What must be Peter's score on the 5th test for his average score on the 5 tests to be 80? \\
        \textbf{Student Process}: 
        \[
        80 \times 5 - 70 \times 4 = 120
        \] \\
        \hline
        \textbf{Error Type:} Careless Error \\
        \textbf{Error Equation:} 
        \[
        80 \times 5 - 70 \times 4 = 120
        \] \\
        \textbf{Teacher Feedback:}
        Carelessly misreading the numbers in the problem, the original average was 78 instead of 70. Therefore, the correct process is 805 - 784 = 88.
        \\
        \bottomrule
    \end{tabular}
    \caption{Example of  Careless Error}
    \label{tab:example_careless}
\end{table}

\noindent\textbf{Lack of Necessary Mathematical Concepts:} 
The ``Lack of Necessary Mathematical Concepts'' category refers to errors that occur when students lack fundamental mathematical knowledge or techniques needed to solve a problem. 
These errors often result in students being unable to attempt the problem at all, as they may not possess the requisite understanding of essential concepts such as fractions, percentages, or algebraic expressions.
Table~\ref{tab:example_lack} presents an example of this type of error. 
The question involved calculating compound interest, but the student responded with ``Do not know how to calculate compound interest.'' indicating a lack of knowledge on how to perform this calculation. As a result, the student was unable to arrive at the correct solution for the problem.

\begin{table}[t]
    \centering
    \footnotesize
    \setlength\tabcolsep{3pt}
    \begin{tabular}{p{7.2cm}}
        \toprule
        \textbf{Problem:}If \$5000 is invested in an account that earns 12\% interest compounded semi-annually, then the interest earned after one year would be how much greater than if the \$5000 had been invested at 8\% simple yearly interest? \\
        \textbf{Student Process}: 
        Do not know how to calculate compound interest. \\
        \hline
        \textbf{Error Type:} Lack of Necessary Mathematical Concepts \\
        \textbf{Error Equation:} 
        None \\
        \textbf{Teacher Feedback:} When calculating compound interest, you also need to consider the interest generated from the previous year's interest. The formula is 
        \[
        A = P \left(1 + \frac{r}{n}\right)^{nt}
        \]
        where \(P\) is the principal, \(r\) is the annual interest rate, \(n\) is the number of times interest is compounded per year, and \(t\) is the number of years the money is invested or borrowed. Therefore, the calculation for compound interest in this question is 
        \[
        A = 5000 \left(1 + 0.12\right)^2.
        \] \\
        \bottomrule
    \end{tabular}
    \caption{Example of Lack of Necessary Mathematical Concepts}
    \label{tab:example_lack}
\end{table}

\section{Student Problem-Solving Records}\label{sec:data_stat}

Table~\ref{tab:student_record} summarizes the correctness of students' solutions together with their problem-solving behaviors (e.g., equations and words per problem), illustrating the diversity of data collected even from only six participants.

\begin{table*}[t]
  \centering
  \scriptsize
    \begin{tabular}{lrrrrrrr|cc}
    \toprule
          & \multicolumn{1}{c}{\textbf{All}} & \multicolumn{1}{c}{\textbf{General}} & \multicolumn{1}{c}{\textbf{Gain}} & \multicolumn{1}{c}{\textbf{Physics}} & \multicolumn{1}{c}{\textbf{Geometry}} & \multicolumn{1}{c}{\textbf{Probability}} & \multicolumn{1}{c|}{\textbf{Other}} & \multicolumn{1}{c}{\textbf{Avg. Equations/Problem}} & \multicolumn{1}{c}{\textbf{Avg. Words/Problem}} \\
    \midrule
    Student 1 & 70.57\% & 67.75\% & 72.73\% & 74.73\% & 69.39\% & 28.57\% & 72.97\% & 3.9516 & 0.6442 \\
    Student 2 & 87.59\% & 86.53\% & 87.22\% & 88.83\% & 87.18\% & 100.00\% & 91.67\% & 7.3124 & 6.7532 \\
    Student 3 & 71.09\% & 70.11\% & 65.44\% & 77.78\% & 64.29\% & 42.86\% & 81.25\% & 5.5958 & 5.0943 \\
    Student 4 & 75.30\% & 77.24\% & 66.43\% & 77.84\% & 78.26\% & 87.50\% & 77.42\% & 6.7212 & 4.0303 \\
    Student 5 & 67.01\% & 66.55\% & 61.15\% & 73.12\% & 65.85\% & 75.00\% & 60.61\% & 4.9237 & 2.8871 \\
    Student 6 & 80.61\% & 77.48\% & 78.74\% & 88.00\% & 72.73\% & 85.71\% & 85.29\% & 4.1484 & 0.3106 \\
    \bottomrule
    \end{tabular}%
  \caption{Correctness and Problem-Solving Characteristics of Students.}
  \label{tab:student_record}%
\end{table*}%

\section{Additional Results on Feedback Generation}
\label{app:rouge12}

Table~\ref{tab:rouge12} reports supplementary results for the feedback generation task using ROUGE-1 and ROUGE-2. 
These metrics reflect n-gram level lexical overlap, which may be less aligned with semantic fidelity and pedagogical usefulness. 
We therefore include them here for completeness, while focusing on ROUGE-L, BERTScore, and LLM Rating in the main text.

\begin{table}[t]
  \centering
  \small
  \setlength\tabcolsep{3mm}
    \begin{tabular}{rlrrr}
    \toprule
          &       & \multicolumn{1}{c}{R-1} & \multicolumn{1}{c}{R-2} & \multicolumn{1}{c}{R-L} \\
    \midrule
    \multicolumn{1}{l}{Llama3 8B} & w/o $r$ & 0.2593 & 0.0809 & 0.2391 \\
          & w/ $r$ & 0.1819 & 0.0557 & 0.1632 \\
    \multicolumn{1}{l}{Llama3 70B} & w/o $r$ & 0.2587 & 0.0881 & 0.2439 \\
          & w/ $r$ & 0.2456 & 0.0876 & 0.2311 \\
    \multicolumn{1}{l}{GPT-3.5} & w/o $r$ & \textbf{0.2897} & 0.1041 & 0.2695 \\
          & w/ $r$ & 0.2848 & 0.1025 & 0.2597 \\
    \multicolumn{1}{l}{o1-mini} & w/o $r$ & 0.1956 & 0.0633 & 0.1839 \\
    \midrule
    \multicolumn{1}{l}{Single-task } & w/o $r$ & 0.1661 & 0.0730 & 0.1594 \\
          & w/ $r$ & 0.0790 & 0.0378 & 0.0750 \\
    \multicolumn{1}{l}{Multi-task } & w/o $r$ & 0.0130 & 0.0049 & 0.0124 \\
          & w/ $r$ & 0.1650 & 0.0719 & 0.1552 \\
    \multicolumn{1}{l}{End-to-end } & w/o $r$ & 0.1862 & 0.0751 & 0.1761 \\
          & w/ $r$ & 0.2842 & \textbf{0.1418} & \textbf{0.2703} \\
    \bottomrule
    \end{tabular}%
  \caption{ROUGE results (R-1, R-2, R-L) for the feedback generation task.}
  \label{tab:rouge12}%
\end{table}%

\section{Evaluation Criteria for Generated Feedback Quality}\label{sec:eval_criteria}

The evaluation criteria are based on a 0 to 3 scale, as follows:
\begin{itemize}[nolistsep]
\item 0: The model's feedback did not help the student at all.
\item 1: The model's feedback pointed in the right direction but contained errors, making it unhelpful to the student.
\item 2: The model's feedback correctly indicated how to correct the student's mistake but lacked clarity, potentially causing the student difficulty in understanding.
\item 3: The model's feedback effectively guided the student on how to correct the mistake, with clear and detailed explanations.
\end{itemize}

\begin{table*}[t]
    \centering
    \footnotesize
    \caption{Comparison of Model-Generated Suggestions on Arithmetic Problem.}
    \begin{tabular}{p{14cm}}
        \toprule
        \textbf{Problem:}maria sold 10 kg of apples in her first hour at the market , but only 2 kg of apples in the second hour . on average , how many kg of apples did she sell in two hours at the market ? \\
        \textbf{Student Process}: 
        10+2=12 \\
        \hline
        \textbf{Human:} After calculating the total number of apples sold in two hours, you still need to divide by the time to get the average sales per hour. Therefore, the answer is 12/2 = 6. \\
        \textbf{Llama3 70B:} The student has calculated the total amount of apples sold, but not the average. To find the average, divide the total amount by the number of hours: 
        \[
        \frac{(10 + 2)}{2} = \frac{12}{2} = 6.
        \] 
        \textbf{GPT-3.5:} 
        To find the average, you need to add the total amount of apples sold in both hours and then divide by the total number of hours. In this case, Maria sold 10 kg in the first hour and 2 kg in the second hour, so the total amount of apples sold in two hours is 10 kg + 2 kg = 12 kg. To find the average, you need to divide this total by 2 (the number of hours), so the average amount of apples sold in two hours is 12 kg / 2 = 6 kg. Therefore, Maria sold an average of 6 kg of apples in two hours at the market. \\
        \textbf{o1-mini:}
        The student correctly calculated the total amount of apples sold by adding the quantities from each hour: 
        \[
        10 \, \text{kg} + 2 \, \text{kg} = 12 \, \text{kg}
        \]
        However, the question asks for the average amount of apples sold per hour, not the total.
        
        To find the average:
        \begin{enumerate}
            \item Total apples sold: 
            \[
            10 \, \text{kg} + 2 \, \text{kg} = 12 \, \text{kg}
            \]
            \item Total time:
            \[
            2 \, \text{hours}
            \]
            \item Average per hour: 
            \[
            \frac{12 \, \text{kg}}{2 \, \text{hours}} = 6 \, \text{kg/hour}
            \]
        \end{enumerate}
        
        Advice: Always pay close attention to whether a question is asking for a total, an average, or another specific calculation. In this case, dividing the total by the number of hours provides the correct average.
        \\
        \textbf{End-to-end:} The question asks for the average number of kilograms of apples sold in two hours, not the total number of kilograms of apples sold in two hours. Therefore, you need to divide the total number of kilograms of apples sold in two hours by 2 to get the average number of kilograms of apples sold in two hours. \\
        \bottomrule
    \end{tabular}
    \label{tab:feedback_ex1}
\end{table*}

\begin{table*}[t]
    \centering
    \footnotesize
    \caption{Comparison of Model-Generated Suggestions to a Geometric Problem Involving Volume Calculation.}
    \begin{tabular}{p{14cm}}
        \toprule
        \textbf{Problem:}a certain rectangular crate measures 8 feet by 12 feet by 14 feet . a cylindrical gas tank is to be made for shipment in the crate and will stand upright when the crate is placed on one of its six faces . what should the radius of the tank be if it is to be of the largest possible volume ? \\
        \textbf{Student Process}: 
        \[
        \frac{8}{2} = 4 \; \text{feet}
        \] \\
        \hline
        \textbf{Human:} The suggestion from the teacher is that simply taking the shortest edge of the crate to calculate the radius is incorrect. It is necessary to consider the different orientations of the crate. Since the gas tank will be placed inside the crate, the radius of the gas tank will depend on the shorter edge of the face on which the crate is placed. By calculating the scenarios for the three different faces, we can determine that placing the gas tank on the 12 feet by 14 feet face will yield the largest volume, calculated as
        \[
        \left( \frac{12}{2} \right) \times \left( \frac{12}{2} \right) \times \pi \times 8
        \] 
        Therefore, the radius is 6. \\
        \textbf{Llama3 70B:} The student seems to be trying to find the radius of the cylinder, but the calculation is incorrect. Since the crate measures 8 feet by 12 feet by 14 feet, the largest possible radius of the cylindrical tank would be the smallest dimension of the crate, which is 8 feet, divided by 2, which is 4 feet. \\
        \textbf{GPT-3.5:} 
         To find the largest possible volume for the cylindrical gas tank that can fit inside the rectangular crate, the radius of the tank should be half of the smaller dimension of the crate. In this case, the radius should be half of 8 feet, which is 4 feet. \\
        \textbf{o1-mini:}
        The student only considered one possible orientation of the crate by taking the dimension 8 feet and dividing it by 2 to get a radius of 4 feet. However, to achieve the largest possible volume for the cylindrical gas tank, it's essential to evaluate all possible orientations of the crate.

        Here's how to approach the problem correctly:
        
        \begin{enumerate}
            \item Identify all orientations:
            \begin{enumerate}
                \item First Orientation: Placing the crate on the 8 ft $\times$ 12 ft face.
                \begin{itemize}
                    \item Height of cylinder: 14 ft
                    \item Maximum diameter: 8 ft (since the cylinder must fit within the 8 ft width)
                    \item Radius: \( \frac{8}{2} = 4 \) ft
                    \item Volume: \( \pi \times 4^2 \times 14 = 224\pi \) cubic feet
                \end{itemize}
        
                \item Second Orientation: Placing the crate on the 12 ft $\times$ 14 ft face.
                \begin{itemize}
                    \item Height of cylinder: 8 ft
                    \item Maximum diameter: 12 ft
                    \item Radius: \( \frac{12}{2} = 6 \) ft
                    \item Volume: \( \pi \times 6^2 \times 8 = 288\pi \) cubic feet
                \end{itemize}
        
                \item Third Orientation: Placing the crate on the 8 ft $\times$ 14 ft face.
                \begin{itemize}
                    \item Height of cylinder: 12 ft
                    \item Maximum diameter: 8 ft
                    \item Radius: \( \frac{8}{2} = 4 \) ft
                    \item Volume: \( \pi \times 4^2 \times 12 = 192\pi \) cubic feet
                \end{itemize}
            \end{enumerate}
        
            \item Determine the optimal orientation:
            \begin{itemize}
                \item The second orientation (placing the crate on the 12 ft $\times$ 14 ft face) provides the largest volume for the cylinder with a radius of 6 feet.
            \end{itemize}
        \end{enumerate}
        
        Conclusion: \\
        To maximize the volume of the cylindrical gas tank, the crate should be placed on the 12 ft $\times$ 14 ft face, allowing the tank to have a radius of 6 feet.
        \\
        \textbf{End-to-end:} The question is misunderstood. The question asks for the radius of the tank, not the length of the crate. \\
        \bottomrule
    \end{tabular}
    \label{tab:feedback_ex2}
\end{table*}

\section{Case Studies of Model-Generated Feedback}\label{sec:Example of Feedback Provided by the Model}

In this section, we present examples of feedback generated by the models to analyze their distinct behaviors.
Table~\ref{tab:feedback_ex1} is a relatively simple math word problem, specifically asking how many apples Maria sells on average per hour over a two-hour period. 
The student only calculated the total number of apples sold during the two hours without determining the average, resulting in an incorrect answer.

Llama3 70B accurately pointed out that the student only calculated the total number of apples sold rather than the average and provided guidance on how to proceed with the calculation, offering excellent advice. 
GPT-3.5 correctly noted that the student calculated the total number of apples sold instead of the average and explained how to perform the subsequent calculations. However, while their suggestions were detailed, the content was overwhelming for the student, making it difficult for them to quickly understand how to correct their answer.
On the other hand, our fine-tuned end-to-end model effectively highlighted that the student needed to divide the total by 2 to arrive at the required average number of apples sold, providing valuable suggestion.

Table~\ref{tab:feedback_ex2} presents another example where the question asks which orientation of a water tank placed in a crate measuring 8 by 12 by 14 feet would yield the largest volume.
The tank is to be placed upright, and the student mistakenly used the smallest dimension divided by 2 as the radius, leading to an error. The problem requires consideration of different orientations of the crate, ultimately determining that placing the crate on the 12 by 14 face yields the maximum tank volume, with the correct radius being 6 feet.
Llama3 70B recognized that the student made an error but ended up performing the same calculation as the student, arriving at the same incorrect answer, which was unhelpful. 
Similarly, GPT-3.5 also failed to consider the varying tank volumes based on different crate orientations. It acknowledged the student's mistake but provided the same incorrect answer, rendering it similarly unhelpful.

In contrast, o1-mini demonstrated the strongest mathematical reasoning abilities, correctly identifying the solution to the problem. It provided a detailed explanation of each step and clarified the reasons for the student's mistake. 
Although the explanation was somewhat lengthy, it still constituted a helpful suggestion. 
The end-to-end model not only failed to offer a correct correction but also misinterpreted the student's process, mistakenly assuming the student was calculating the crate's dimensions. 
This model performed the worst in this instance.

\section{GenAI Usage Disclosure}

We utilized generative AI tools solely for grammar and language refinement. 
All content was reviewed and edited by the author(s), who take full responsibility for the final manuscript.

\end{document}